\documentclass[11pt,a4paper]{article}
\usepackage[hyperref]{acl2019}
\usepackage{times}
\usepackage{latexsym}
\usepackage[draft]{changes}

\definechangesauthor[color=blue]{mm} 
\definechangesauthor[color=red]{todo}
\definechangesauthor[color=orange]{notes}

\usepackage{url}

\aclfinalcopy 
\def\aclpaperid{2418} 

\title{Implicit Discourse Relation Identification for Open-domain Dialogues}
 
\author{
    Mingyu Derek Ma\textsuperscript{1}, Kevin K. Bowden\textsuperscript{2}, Jiaqi Wu\textsuperscript{2}, Wen Cui\textsuperscript{2} \and Marilyn Walker\textsuperscript{2}\\
    \textsuperscript{1}Human-Computer Communications Laboratory \&\\ Stanley Ho Big Data Decision Analytics Research Centre\\ The Chinese University of Hong Kong\\
    {\tt derekma@cuhk.edu.hk}\\
  \textsuperscript{2}Natural Language and Dialogue Systems Lab\\
  University of California, Santa Cruz\\
  {\tt \{kkbowden, jwu64, wcui7, mawalker\}@ucsc.edu}
}

\date{}
\begin{document}
\maketitle
\begin{abstract}
    Discourse relation identification has been an active area of research for many years, and the challenge of identifying implicit relations remains largely an unsolved task, especially in the context of an open-domain dialogue system. Previous work primarily relies on a corpora of formal text which is inherently non-dialogic, i.e., news and journals. This data however is not suitable to handle the nuances of informal dialogue nor is it capable of navigating the plethora of valid topics present in open-domain dialogue. In this paper, we designed a novel discourse relation identification pipeline specifically tuned for open-domain dialogue systems. We firstly propose a method to automatically extract the implicit discourse relation argument pairs and labels from a dataset of dialogic turns, resulting in a novel corpus of discourse relation pairs; the first of its kind to attempt to identify the discourse relations connecting the dialogic turns in open-domain discourse. Moreover, we have taken the first steps to leverage the dialogue features unique to our task to further improve the identification of such relations by performing feature ablation and incorporating dialogue features to enhance the state-of-the-art model.
\end{abstract}

\section{Introduction}
Discourse analysis considering relations between clauses has received increasing attention from the field, and implicit discourse relation identification is one of the most challenging problems in discourse parsing since it is purely based on textual features. Previous work has defined four widely accepted major classes of discourse relation - ``Comparison'', ``Expansion'', ``Contingency'' and ``Temporal'' \cite{miltsakaki2008sense, prasad2018thepenn}. These four relations can either be explicitly or implicitly realized. When explicitly realized, there are often clear connective words between clauses which result in an associated discourse relation, while implicit realizations are often much harder to detect. For example, people can imply there is a ``Comparison'' relation between the following two sentences by understanding the meaning. Without clear keywords like ``but'' however, it is hard for machines to recognize such implicit relations.\\
\indent\indent\indent{Arg 1: \textit{it's a great album.}} \\
\indent\indent\indent{Arg 2: \textit{it's probably not their best.}}

Since the development of the Penn Discourse Treebank (PDTB)\footnote{More details about Penn Discourse Treebank can be found at \url{https://www.seas.upenn.edu/~pdtb/}}, discourse relation identification has been treated as a supervised learning problem. For explicit discourse relation pairs, simple classification methods based on connective cues achieve more than 90\% accuracy \cite{pitler2008easily}. For implicit discourse relations however, where there is no discourse clue, relations needs to be inferred on the basis of textual features, making this a challenging  problem in discourse parsing \cite{li2014addressing,lin2009recognizing}.

While previous work has suggested that discourse relations may hold between dialogue turns, this idea is relatively unexplored \cite{stent2000rhetorical,tonelli2010annotation}. We posit that discourse relation identification could have wide application in dialogue systems,
by cultivating a more aware state space in order to improve the continuity between an extended sequence of turns.  The detected discourse relation could additionally serve as a query or ranking parameter for possible next turns, retrieved from a database of content, or generated by natural language generation. Adding this additional natural language understanding component might be especially useful when navigating open-domain dialogue where user input is  unpredictable and the model must be topic-robust.

There are many fundamental challenges with identifying and utilizing discourse relations in an open-domain dialogue system. All existing datasets for discourse relation identification are based on monologic text such as news; these datasets are unlikely to provide good training material for dialogue. Moreover there is no previous work  investigating the feasibility of applying a machine learning model developed on formal text to dialogic content, where turns in are normally short, informal text. Thus, the lack of labeled dialogue data for implicit discourse relation pairs in  open-domain dialogue is the first challenge that must be addressed.


To tackle these two problems and utilize the unexplored benefits of features unique to dialogue systems, we carry out two steps. First, we construct a discourse relation pair dataset from a large corpus of open-domain dialogue, which to our knowledge is the first of its kind. Second, we investigated a feature-based model with different dialogue feature combinations and enhanced a deep learning model by incorporating dialogue features that utilize aspects unique to dialogue. The dataset and related code are publicly available.\footnote{\href{https://github.com/derekmma/dialogue-discourse-relation}{{\tt https://github.com/derekmma/}}\\\href{https://github.com/derekmma/dialogue-discourse-relation}{{\tt dialogue-discourse-relation}}}

\section{Related Work}
The release of the Penn Discourse Treebank (PDTB) \cite{prasad2018thepenn} makes research on machine learning based implicit discourse relation recognition possible. Most previous work is based on linguistic and semantic features such as word pairs and brown cluster pair representation \cite{pitler2008easily, lin2009recognizing} or rule-based systems \cite{wellner2006classification}. Recent work has proposed neural network based models with attention or advanced representations, such as CNN \cite{qin2016stacking}, attention on neural tensor network \cite{guo2018implicit}, and memory networks \cite{jia2018modeling}. Advanced representations may help to achieve higher performance \cite{bai2018deep}. Some methods also consider context paragraphs and inter-paragraph dependency \cite{dai2018improving}.

To utilize machine learning models for this task, larger datasets would provide a bigger optimization space \cite{li2014addressing}. \citet{marcu2002unsupervised} is the first work to generate artificial samples to extend the dataset by using rules to convert explicit discourse relation pairs into implicit pairs by dropping the connectives. This work is further extended by methods for selecting high-quality samples  \cite{rutherford2015improving, xu2018using, braud2014combining, wang2012implicit}.

Most of the existing work discussed so far is based on the PDTB dataset, which targets formal texts like news, making it less suitable for our task which is centered around informal dialogue. Related work on discourse relation annotation in a dialogue corpus is limited \cite{stent2000rhetorical,tonelli2010annotation}. For example \citet{tonelli2010annotation} annotated the Luna corpus,\footnote{EU FP6 contract No. 33549, \url{http://www.ist-luna.eu/}}  which does not include English annotations. To our knowledge there is no English dialogue-based corpus with implicit discourse relation labels, as such research specifically targeting a discourse relation identification model for social open-domain dialogue remains unexplored.

\section{Dataset Construction}

Previous work on discourse relation identification suggests that the most effective approach is supervised learning, but limited amounts of annotated data constrain the application of such algorithms. Previous work has additionally proven that weakly labeled data, which contains a small number of false labels and can be generated automatically, helps  improve  classifier performance with implicit relations \cite{rutherford2015improving}. 

We therefore constructed Edina-DR, the novel dataset of discourse relation pairs based on the publicly available self-dialogue Edina corpus which contains 24,165 multi-turn social conversations across 23 topics \citep{fainberg2018talking, krause2017edina}.\footnote{The Edina dataset is publicly available at \href{https://github.com/jfainberg/self_dialogue_corpus}{{\tt https://github.com/jfainberg/self\underline{ }}}\\\href{https://github.com/jfainberg/self_dialogue_corpus}{{\tt dialogue\underline{ }corpus}}} To the best of our knowledge, this is the first English discourse relation dataset based on open-domain dialogues. The Edina dataset initially contains no discourse relation labels. Inspired by the approaches taken to automatically extend PDTB, we designed a pipeline to extract discourse relation argument pairs through utilizing the connective words which are known as clear relation indicators.  The pipeline  automatically extracts argument pairs and assign discourse relation labels to each of the utterances. We then have humans annotate a small sample of the data in order to validate the automated pipeline. Our  pipeline targets the four level-1 discourse relations, i.e., ``Comparison'', ``Expansion'', ``Contingency'' and ``Temporal''.

We obtained this initial connectives pool according to statistical analysis of connective frequencies in PDTB conducted by \citet{pitler2008easily}, in which we only consider connectives which are strongly associated (probability $>$ 95\%) with only one class of relation.\footnote{The list of connectives for each relation in detail can be found in \cite{pitler2008easily}.} For example, we exclude the connective word ``since'' because it may often appear as an indicator of either a ``Temporal'' or ``Contingency'' relation. 

Secondly, some connectives cannot be removed without changing the original meaning \cite{sporleder2008using}. We follow the method proposed by \citet{rutherford2015improving} to identify the connectives which are freely omissible by measuring the Omissible Rate and Context Differential. Since we need some manually labeled connectives for this task, we implement the connective selection on the PDTB dataset and generalize the selection result to the dialogue dataset. The selected connectives include:
\begin{itemize}
    \item \textbf{Comparison}: but, however, although, by contrast
    \item \textbf{Contingency}: because, so, thus, as a result, consequently, therefore
    \item \textbf{Expansion}: also, for example, in addition, instead, indeed, moreover, for instance, in fact, furthermore, or, and
    \item \textbf{Temporal}: then, previously, earlier, later, after, before
\end{itemize}

The third step is to select the conversations matching specific predefined patterns for different structures of the sentences with the selected connective words shown above. Inspired by \cite{braud2014combining, marcu2002unsupervised}, we use two patterns: {\tt (Arg 1) (connective) (Arg 2)} and {\tt (Arg 1). (Connective),(Arg 2)}. In other words, we have one pattern for when connectives appear in the middle of an utterance, and another pattern for when connectives link two arguments in adjacent utterances across separate turns. Finally, we defined several heuristic rules to filter out low-quality pairs which have been applied in previous work \cite{braud2014combining}. The program only accepts full sentence arguments and we use certain POS tags for particular connectives to make sure the connective function as relation indicators. A segment window is defined so that our method only picks the closest phrases or sub-sentences if the whole conversation contains several sentences.


For example, in the sentence ``\textit{they had a \$5 off the price, \textbf{so} i bought it.}'', the connective ``so'' is identified in the list of connective words for ``Contingency'' relation and the sentence matches our pattern 1. Therefore we convert this sentence to a ``Contingency'' discourse relation pair and the two arguments are ``they had a \$5 off the price'' and ``i bought it''.

\begin{table}[h]
    \centering
    \begin{tabular}{l|r|r}
    \hline\hline
      & Edina-DR & PDTB \\\hline
    \# pairs of all relations & 27998 & 11734 \\
    avg \# words of arg 1 & 7.1 & 18.8\\
    avg \# words of arg 2 & 7.3 & 19.4\\
    \hline
    \# pairs of `Comparison' & 20823 & 1799\\
    \# pairs of `Contingency' & 5080 & 2243\\
    \# pairs of `Expansion' & 1580 & 6933\\
    \# pairs of `Temporal' & 452 & 759\\
    \hline\hline
    \end{tabular}
    \caption{Statistics of the extracted dataset Edina-DR}
    \label{tab:stat_datasets}
\end{table}

The statistics of the annotated dialogue discourse relation pairs dataset Edina-DR is shown in Table \ref{tab:stat_datasets}. The new dataset contains more than twice the pairs compared to PDTB, which should prove useful for machine learning. We note that the distribution of discourse relations in the Edina-DR dataset is different from PDTB. Most of the pairs belong to the ``Comparison'' relation, which is a natural way to structure dialogue. The number of ``Temporal'' pairs however is smaller, one possible explanation being that people do not use connectives words often in dialogues when talking about time-related events. These differences highlight the need for this work, as it's clear that human dialogue is in fact structured differently than more formal non-dialogic text. 

We annotated discourse relations for 400 samples out of the extracted dataset by an expert annotator, 12\% of the samples do not form a discourse relation which probably due to failures  by  the automatic extraction program  to catch particular linguistic structures. 88\% of the samples which do hold relations match the relation labels of the human annotations, which proves the reliability of our proposed extraction method.

\section{Model}
We propose the novel approach of applying the unique dialogue features encapsulated in the state-space of a real deployed dialogue systems to enhance discourse relation identification. Firstly, we use a feature-based classifier for feature selection and then we explore the feasibility of utilizing existing deep learning model in dialogue discourse relation identification task.

\subsection{Feature-based Classifier}
    We extract dialogue features using the Natural Language Understanding (NLU) capabilities in SlugBot, a deployed open-domain dialogue system \cite{bowdenslugbot, bowden2018slugbot}. These features are normally used for dialogue management and content retrieval. We input raw argument pairs into the NLU pipeline and get dialogue features which are then fed as one-hot vectors to a logistic regression classifier. A full dialogue feature vector contains 448 features. The dialogue features include:\\
        \textbf{Dialogue Act}: The act of a dialogue utterance is obtained using the NPS dialogue act classifier \cite{forsythand2007lexical}. There are 15 different dialogue acts, including \textsc{Greet}, \textsc{Clarify}, and \textsc{Statement}. The full list of dialogue acts is described in \cite{forsythand2007lexical}.\\
        \textbf{Sentiment}: The sentiment of a dialogue utterance is obtained from the Stanford CoreNLP Toolkit \cite{manning2014stanford} and there are five possible sentiment values: \textsc{very positive}, \textsc{positive}, \textsc{neutral}, \textsc{negative}, and \textsc{very negative}. \\
        \textbf{Intent}: An utterance intent ontology consisting of 33 discrete intents is developed and recognized using heuristics and a trained model. It is designed to obtain utterance intent without conversational context, so only the input utterance is considered for intent detection. Some sample intents are \textsc{request\_opinion}, \textsc{request\_service}, \textsc{request\_change\_topic}. It is trained using a subset of Common Alexa Prize Chats (CAPC) dataset with roughly 50K utterances and the model ensembles both a Recurrent Neural Network and Convolutional Neural Network \cite{ram2018conversational}.\\
        \textbf{Topic}: The topic of the utterance is obtained using the CoBot (Conversational Bot) toolkit topic classification model \cite{khatri2018advancing}, which is a Deep Average network BiLSTM model. The model is trained on over 120,000 utterances and labeled across 22 topics. This includes commonly discussed topics such as \textsc{politics}, \textsc{fashion}, \textsc{sports}, \textsc{science and technology}, and \textsc{music}.\\
        \textbf{Core Entities Types}: We use SlugNERDS to detect our named entities \cite{bowden2018slugnerds, bowden2017combining}. SlugNERDS is specialized for open-domain dialogue interactions. It can sift through noisy user data and it uses the constantly updated Google Knowledge Graph\footnote{\href{https://developers.google.com/knowledge-graph/}{{\tt https://developers.google.com/}}\\\href{https://developers.google.com/knowledge-graph/}{{\tt knowledge-graph/}}} to remain aware of even the latest named entities. Both of these points are vital for understanding social chit-chat. We only consider the entity types of the entities as feature rather than entities themselves. We use standard \url{schema.org} types and there are totally 614 types. For example, if SlugNERDS detects ``Cam Newton'', which is an entity with type \textsc{person}, then \textsc{person} is used as feature.

\subsection{Deep Learning Model with Dialogue Features}
To investigate the adaptability of existing discourse relation identification models on dialogue data and our proposed features, we build on the Deep Enhanced Representation (DER) model of \citet{bai2018deep}\footnote{Original implementation of the authors can be found at \href{https://github.com/hxbai/Deep_Enhanced_Repr_for_IDRR}{{\tt https://github.com/hxbai/Deep\underline{ }Enhanced\underline{ }}}\\\href{https://github.com/hxbai/Deep_Enhanced_Repr_for_IDRR}{{\tt Repr\underline{ }for\underline{ }IDRR}}.}, which demonstrated its efficiency by achieving the current state-of-the-art performance on the PDTB dataset. It utilized different grained text representations including character, sub-word, word, sentence, and sentence pair levels, with embeddings obtained by ELMo \cite{Peters:2018}. The model first generates representations for the argument pairs using an encoder and bi-attention module; these are then sent to the classifier, consisting of multiple layer perceptrons with softmax, to predict the discourse relation.

We take the DER design and architecture and train on Edina-DR dataset to evaluate the adaptability of existing model in dialogue environment. Then we explore a variation of this model by connecting dialogue feature vectors to the argument pairs representation vector to extend the representation. We use the same method to encode all dialogue features as the feature-based classifier. With the help of previous experiments, we use the best feature combination for the dialogue feature vectors.

\section{Evaluation and Analysis}
For the following experiments, we randomly selected 400 samples to be used as test set with discourse relation labels annotated by an expert. We repeat the experiments  five times and take the average score as the final report results.
\subsection{Feature-based Classifier and Dialogue Feature Selection}
We first analyze the performance of the feature-based model with different feature combinations shown in Table \ref{tab:experiment_1}.

\begin{table}[h]
    \centering
    \begin{tabular}{l|r|r|r}
    \hline\hline
    Features & Precision & Recall & F1  \\ \hline
    \textsc{dialogue act} & 0.64 & 0.69 & 0.66 \\ 
    \textsc{intent} & 0.63 & 0.74 & 0.68 \\ 
    \textsc{topics} & 0.62 & 0.71 & 0.66 \\ 
    \textsc{sentiment} & 0.56 & 0.74 & 0.64 \\ 
    \textsc{entities types} & 0.63 & 0.74 & 0.68 \\ \hline
    All & 0.63 & 0.65 & 0.64 \\ 
    All - \textsc{sentiment} & 0.64 & 0.73 & 0.68 \\
    \hline\hline
    \end{tabular}
    \caption{Feature-based Model Evaluation}
    \label{tab:experiment_1}
\end{table}

For single dialogue features, \textsc{intent} and \textsc{entities types} provide the largest performance boost compared to other single dialogue features, and this demonstrates the effectiveness of using intent and types of entities for discourse relation identification. Other three features maintain the same level of performance, except a large drop in precision with respect to \textsc{sentiment}. One possible explanation is that our sentiment classification results are obtained using the Sentiment Annotator from Stanford CoreNLP Toolkit, which is trained on movie reviews corpus \cite{manning2014stanford, socher2013recursive}. The nature of training data is not suitable for our dialogue corpus in this task. Using Table \ref{tab:experiment_1}, we see that the best configuration includes all of our dialogue features except \textsc{sentiment}.

\subsection{Deep Learning Models}


\begin{table}[h]
    \centering
    \begin{tabular}{l|r|r}
    \hline\hline
    Model & Acc. & F1  \\ \hline
    DER (PDTB) & 0.61 & 0.51 \\
    Logistic Reg. (Edina-DR) & 0.64 & 0.68  \\
    DER (Edina-DR) & 0.80 & 0.76 \\
    DER+Dialogue (Edina-DR) & \textbf{0.81} & \textbf{0.77} \\
    
    \hline\hline
    \end{tabular}
    \caption{Performance of Deep Learning Models (Dataset name is shown in parentheses)}
    \label{tab:experiments_2}
\end{table}

In Table \ref{tab:experiments_2}, we see the results of our experiments, where  DER represents our baseline model. We use the default parameter for DER models. We also show the result of the DER model trained and tested on the PDTB dataset for comparison marked as ``DER (PDTB)''. The first observation is that the DER model performs surprisingly well with an F1 score of 0.76 on the new dialogue discourse relation dataset Edina-DR with p-value of 0.008, which demonstrates its strong adaptability to the task of discourse relation identification in dialogues. Comparing the same DER model on PDTB, the large drop in F1 score shows the difference between formal and informal data. We also find that the model with dialogue features enhance the performance by 1\% on F1 score with p-value 0.006, which indicates the potential of using dialogue features to further enhance discourse relation identification models.


\section{Conclusion and Future Work}
In this paper, we proposed a novel pipeline specifically designed for implicit discourse relation identification in open-domain dialogue. We constructed a novel dataset of discourse relation pairs for dialogue conversations, and utilized unique dialogue features to enhance the performance of a state-of-the-art classifier. Our experiments show that dialogue intent and entities types play important roles and dialogue features can increase the performance of the discourse relation identification model. 

Since implicit discourse relation identification is a key task for dialogue systems, there are still many approaches worth investigating in future work. More sophisticated dialogue features and classification algorithms are needed for the discourse relation identification task in addition to a larger more balanced corpus.


\bibliography{acl2019}

\begin{thebibliography}{31}
\expandafter\ifx\csname natexlab\endcsname\relax\def\natexlab#1{#1}\fi

\bibitem[{Bai and Zhao(2018)}]{bai2018deep}
Hongxiao Bai and Hai Zhao. 2018.
\newblock Deep enhanced representation for implicit discourse relation
  recognition.
\newblock In \emph{Proceedings of the 27th International Conference on
  Computational Linguistics}, pages 571--583.

\bibitem[{Bowden et~al.(2017)Bowden, Oraby, Wu, Misra, and
  Walker}]{bowden2017combining}
Kevin~K Bowden, Shereen Oraby, Jiaqi Wu, Amita Misra, and Marilyn Walker. 2017.
\newblock Combining search with structured data to create a more engaging user
  experience in open domain dialogue.
\newblock \emph{arXiv preprint arXiv:1709.05411}.

\bibitem[{Bowden et~al.()Bowden, Wu, Cui, Juraska, Harrison, Schwarzmann,
  Santer, and Walker}]{bowdenslugbot}
Kevin~K Bowden, Jiaqi Wu, Wen Cui, Juraj Juraska, Vrindavan Harrison, Brian
  Schwarzmann, Nick Santer, and Marilyn Walker.
\newblock Slugbot: Developing a computational model and framework of a novel
  dialogue genre.

\bibitem[{Bowden et~al.(2018{\natexlab{a}})Bowden, Wu, Oraby, Misra, and
  Walker}]{bowden2018slugbot}
Kevin~K Bowden, Jiaqi Wu, Shereen Oraby, Amita Misra, and Marilyn Walker.
  2018{\natexlab{a}}.
\newblock Slugbot: An application of a novel and scalable open domain socialbot
  framework.
\newblock \emph{arXiv preprint arXiv:1801.01531}.

\bibitem[{Bowden et~al.(2018{\natexlab{b}})Bowden, Wu, Oraby, Misra, and
  Walker}]{bowden2018slugnerds}
Kevin~K Bowden, Jiaqi Wu, Shereen Oraby, Amita Misra, and Marilyn Walker.
  2018{\natexlab{b}}.
\newblock Slugnerds: A named entity recognition tool for open domain dialogue
  systems.
\newblock \emph{arXiv preprint arXiv:1805.03784}.

\bibitem[{Braud and Denis(2014)}]{braud2014combining}
Chlo{\'e} Braud and Pascal Denis. 2014.
\newblock Combining natural and artificial examples to improve implicit
  discourse relation identification.
\newblock In \emph{Proceedings of COLING 2014, the 25th International
  Conference on Computational Linguistics: Technical Papers}, pages 1694--1705.

\bibitem[{Dai and Huang(2018)}]{dai2018improving}
Zeyu Dai and Ruihong Huang. 2018.
\newblock Improving implicit discourse relation classification by modeling
  inter-dependencies of discourse units in a paragraph.
\newblock In \emph{Proceedings of the 2018 Conference of the North American
  Chapter of the Association for Computational Linguistics: Human Language
  Technologies, Volume 1 (Long Papers)}, volume~1, pages 141--151.

\bibitem[{Fainberg et~al.(2018)Fainberg, Krause, Dobre, Damonte, Kahembwe,
  Duma, Webber, and Fancellu}]{fainberg2018talking}
Joachim Fainberg, Ben Krause, Mihai Dobre, Marco Damonte, Emmanuel Kahembwe,
  Daniel Duma, Bonnie Webber, and Federico Fancellu. 2018.
\newblock Talking to myself: self-dialogues as data for conversational agents.
\newblock \emph{arXiv preprint arXiv:1809.06641}.

\bibitem[{Forsyth and Martell(2007)}]{forsythand2007lexical}
Eric~N Forsyth and Craig~H Martell. 2007.
\newblock Lexical and discourse analysis of online chat dialog.
\newblock In \emph{International Conference on Semantic Computing (ICSC 2007)},
  pages 19--26. IEEE.

\bibitem[{Guo et~al.(2018)Guo, He, Jin, Dang, Wang, and Li}]{guo2018implicit}
Fengyu Guo, Ruifang He, Di~Jin, Jianwu Dang, Longbiao Wang, and Xiangang Li.
  2018.
\newblock Implicit discourse relation recognition using neural tensor network
  with interactive attention and sparse learning.
\newblock In \emph{Proceedings of the 27th International Conference on
  Computational Linguistics}, pages 547--558.

\bibitem[{Jia et~al.(2018)Jia, Ye, Feng, Lai, Yan, and Zhao}]{jia2018modeling}
Yanyan Jia, Yuan Ye, Yansong Feng, Yuxuan Lai, Rui Yan, and Dongyan Zhao. 2018.
\newblock Modeling discourse cohesion for discourse parsing via memory network.
\newblock In \emph{Proceedings of the 56th Annual Meeting of the Association
  for Computational Linguistics (Volume 2: Short Papers)}, volume~2, pages
  438--443.

\bibitem[{Khatri et~al.(2018)Khatri, Hedayatnia, Venkatesh, Nunn, Pan, Liu,
  Song, Gottardi, Kwatra, Pancholi et~al.}]{khatri2018advancing}
Chandra Khatri, Behnam Hedayatnia, Anu Venkatesh, Jeff Nunn, Yi~Pan, Qing Liu,
  Han Song, Anna Gottardi, Sanjeev Kwatra, Sanju Pancholi, et~al. 2018.
\newblock Advancing the state of the art in open domain dialog systems through
  the alexa prize.
\newblock \emph{arXiv preprint arXiv:1812.10757}.

\bibitem[{Krause et~al.(2017)Krause, Damonte, Dobre, Duma, Fainberg, Fancellu,
  Kahembwe, Cheng, and Webber}]{krause2017edina}
Ben Krause, Marco Damonte, Mihai Dobre, Daniel Duma, Joachim Fainberg, Federico
  Fancellu, Emmanuel Kahembwe, Jianpeng Cheng, and Bonnie Webber. 2017.
\newblock Edina: Building an open domain socialbot with self-dialogues.
\newblock \emph{arXiv preprint arXiv:1709.09816}.

\bibitem[{Li and Nenkova(2014)}]{li2014addressing}
Junyi~Jessy Li and Ani Nenkova. 2014.
\newblock Addressing class imbalance for improved recognition of implicit
  discourse relations.
\newblock In \emph{Proceedings of the 15th Annual Meeting of the Special
  Interest Group on Discourse and Dialogue (SIGDIAL)}, pages 142--150.

\bibitem[{Lin et~al.(2009)Lin, Kan, and Ng}]{lin2009recognizing}
Ziheng Lin, Min-Yen Kan, and Hwee~Tou Ng. 2009.
\newblock Recognizing implicit discourse relations in the penn discourse
  treebank.
\newblock In \emph{Proceedings of the 2009 Conference on Empirical Methods in
  Natural Language Processing: Volume 1-Volume 1}, pages 343--351. Association
  for Computational Linguistics.

\bibitem[{Manning et~al.(2014)Manning, Surdeanu, Bauer, Finkel, Bethard, and
  McClosky}]{manning2014stanford}
Christopher Manning, Mihai Surdeanu, John Bauer, Jenny Finkel, Steven Bethard,
  and David McClosky. 2014.
\newblock The stanford corenlp natural language processing toolkit.
\newblock In \emph{Proceedings of 52nd annual meeting of the association for
  computational linguistics: system demonstrations}, pages 55--60.

\bibitem[{Marcu and Echihabi(2002)}]{marcu2002unsupervised}
Daniel Marcu and Abdessamad Echihabi. 2002.
\newblock An unsupervised approach to recognizing discourse relations.
\newblock In \emph{Proceedings of the 40th annual meeting of the association
  for computational linguistics}.

\bibitem[{Miltsakaki et~al.(2008)Miltsakaki, Robaldo, Lee, and
  Joshi}]{miltsakaki2008sense}
Eleni Miltsakaki, Livio Robaldo, Alan Lee, and Aravind Joshi. 2008.
\newblock Sense annotation in the penn discourse treebank.
\newblock In \emph{International Conference on Intelligent Text Processing and
  Computational Linguistics}, pages 275--286. Springer.

\bibitem[{Peters et~al.(2018)Peters, Neumann, Iyyer, Gardner, Clark, Lee, and
  Zettlemoyer}]{Peters:2018}
Matthew~E. Peters, Mark Neumann, Mohit Iyyer, Matt Gardner, Christopher Clark,
  Kenton Lee, and Luke Zettlemoyer. 2018.
\newblock Deep contextualized word representations.
\newblock In \emph{Proc. of NAACL}.

\bibitem[{Pitler et~al.(2008)Pitler, Raghupathy, Mehta, Nenkova, Lee, and
  Joshi}]{pitler2008easily}
Emily Pitler, Mridhula Raghupathy, Hena Mehta, Ani Nenkova, Alan Lee, and
  Aravind~K Joshi. 2008.
\newblock Easily identifiable discourse relations.
\newblock \emph{Technical Reports (CIS)}, page 884.

\bibitem[{Prasad et~al.(2008)Prasad, Dinesh, Lee, Miltsakaki, Robaldo, Joshi,
  and Webber}]{prasad2018thepenn}
Rashmi Prasad, Nikhil Dinesh, Alan Lee, Eleni Miltsakaki, Livio Robaldo,
  Aravind Joshi, and Bonnie Webber. 2008.
\newblock The penn discourse treebank 2.0.
\newblock In \emph{Proceedings of the 6th International Conference on Language
  Resources and Evaluation}.

\bibitem[{Qin et~al.(2016)Qin, Zhang, and Zhao}]{qin2016stacking}
Lianhui Qin, Zhisong Zhang, and Hai Zhao. 2016.
\newblock A stacking gated neural architecture for implicit discourse relation
  classification.
\newblock In \emph{Proceedings of the 2016 Conference on Empirical Methods in
  Natural Language Processing}, pages 2263--2270.

\bibitem[{Ram et~al.(2018)Ram, Prasad, Khatri, Venkatesh, Gabriel, Liu, Nunn,
  Hedayatnia, Cheng, Nagar et~al.}]{ram2018conversational}
Ashwin Ram, Rohit Prasad, Chandra Khatri, Anu Venkatesh, Raefer Gabriel, Qing
  Liu, Jeff Nunn, Behnam Hedayatnia, Ming Cheng, Ashish Nagar, et~al. 2018.
\newblock Conversational ai: The science behind the alexa prize.
\newblock \emph{arXiv preprint arXiv:1801.03604}.

\bibitem[{Rutherford and Xue(2015)}]{rutherford2015improving}
Attapol Rutherford and Nianwen Xue. 2015.
\newblock Improving the inference of implicit discourse relations via
  classifying explicit discourse connectives.
\newblock In \emph{Proceedings of the 2015 Conference of the North American
  Chapter of the Association for Computational Linguistics: Human Language
  Technologies}, pages 799--808.

\bibitem[{Socher et~al.(2013)Socher, Perelygin, Wu, Chuang, Manning, Ng, and
  Potts}]{socher2013recursive}
Richard Socher, Alex Perelygin, Jean Wu, Jason Chuang, Christopher~D Manning,
  Andrew Ng, and Christopher Potts. 2013.
\newblock Recursive deep models for semantic compositionality over a sentiment
  treebank.
\newblock In \emph{Proceedings of the 2013 conference on empirical methods in
  natural language processing}, pages 1631--1642.

\bibitem[{Sporleder and Lascarides(2008)}]{sporleder2008using}
Caroline Sporleder and Alex Lascarides. 2008.
\newblock Using automatically labelled examples to classify rhetorical
  relations: An assessment.
\newblock \emph{Natural Language Engineering}, 14(3):369--416.

\bibitem[{Stent(2000)}]{stent2000rhetorical}
Amanda Stent. 2000.
\newblock Rhetorical structure in dialog.
\newblock In \emph{INLG'2000 Proceedings of the First International Conference
  on Natural Language Generation}.

\bibitem[{Tonelli et~al.(2010)Tonelli, Riccardi, Prasad, and
  Joshi}]{tonelli2010annotation}
Sara Tonelli, Giuseppe Riccardi, Rashmi Prasad, and Aravind~K Joshi. 2010.
\newblock Annotation of discourse relations for conversational spoken dialogs.
\newblock In \emph{LREC}.

\bibitem[{Wang et~al.(2012)Wang, Li, Li, and Li}]{wang2012implicit}
Xun Wang, Sujian Li, Jiwei Li, and Wenjie Li. 2012.
\newblock Implicit discourse relation recognition by selecting typical training
  examples.
\newblock \emph{Proceedings of COLING 2012}, pages 2757--2772.

\bibitem[{Wellner et~al.(2006)Wellner, Pustejovsky, Havasi, Rumshisky, and
  Sauri}]{wellner2006classification}
Ben Wellner, James Pustejovsky, Catherine Havasi, Anna Rumshisky, and Roser
  Sauri. 2006.
\newblock Classification of discourse coherence relations: An exploratory study
  using multiple knowledge sources.
\newblock In \emph{Proceedings of the 7th SIGdial Workshop on Discourse and
  Dialogue}, pages 117--125.

\bibitem[{Xu et~al.(2018)Xu, Hong, Ruan, Yao, Zhang, and Zhou}]{xu2018using}
Yang Xu, Yu~Hong, Huibin Ruan, Jianmin Yao, Min Zhang, and Guodong Zhou. 2018.
\newblock Using active learning to expand training data for implicit discourse
  relation recognition.
\newblock In \emph{Proceedings of the 2018 Conference on Empirical Methods in
  Natural Language Processing}, pages 725--731.

\end{thebibliography}
\bibliographystyle{acl_natbib}




\end{document}